\documentclass[10pt,twocolumn,letterpaper]{article}
\usepackage{cvpr}              

\usepackage{multirow}
\usepackage{bbm}

\newcommand{\parahead}[1]{\noindent\textbf{#1}.\ }

\definecolor{r1color}{HTML}{0073C2} 
\definecolor{r2color}{HTML}{D55E00} 
\definecolor{r3color}{HTML}{009E73} 

\definecolor{cvprblue}{rgb}{0.21,0.49,0.74}
\usepackage[pagebackref,breaklinks,colorlinks,allcolors=cvprblue]{hyperref}

\def\paperID{4} 
\def\confName{CV4Animals Workshop}
\def\confYear{2025}

\title{Animal Pose Labeling Using General-Purpose Point Trackers}
\author{
Zhuoyang Pan$^{\dagger, \ddagger}$ \quad
Boxiao Pan$^\dagger$ \quad
Guandao Yang$^\dagger$ \quad
Adam W. Harley$^\dagger$ \quad
Leonidas Guibas$^\dagger$ 
\\[0.5em]
$^\dagger$Stanford University \quad
$^\ddagger$ShanghaiTech University
\vspace{-1em}
}

\begin{document}

\maketitle
\begin{abstract}
Automatically estimating animal poses from videos is important for studying animal behaviors. Existing methods do not perform reliably since they are trained on datasets that are not comprehensive enough to capture all necessary animal behaviors. However, it is very challenging to collect such datasets due to the large variations in animal morphology. In this paper, we propose an animal pose labeling pipeline that follows a different strategy, \ie test time optimization. Given a video, we fine-tune a lightweight appearance embedding inside a pre-trained general-purpose point tracker on a sparse set of annotated frames. These annotations can be obtained from human labelers or off-the-shelf pose detectors. The fine-tuned model is then applied to the rest of the frames for automatic labeling. Our method achieves state-of-the-art performance at a reasonable annotation cost. We believe our pipeline offers a valuable tool for the automatic quantification of animal behavior. Visit our project webpage at \url{https://zhuoyang-pan.github.io/animal-labeling}.
\end{abstract}

\vspace{-2em}
\section{Introduction}
\label{sec:intro}

Accurate quantification of animal behavior is essential to understanding their brain by connecting it to the studied subjects' neural activities~\cite{DeepLabCut2018}. Such quantification requires precise detection and tracking of animal poses, preferably in a markerless fashion to avoid intrusiveness. This necessitates the development of automatic pose detectors and trackers that are applicable to in-the-wild videos.
Despite significant progress in human pose estimation~\cite{OpenPose2019,ViTPose2022,HMR22023}, accurately estimating animal poses remains challenging due to the large variations in body morphology. These variations make it hard to collect comprehensive datasets and thus train generalizable models.

Previous works on animal pose estimation, therefore, focus primarily on specific animal species~\cite{DeepLabCut2018,DeepPoseKit2019,CNNAnimal2019,DeepLabCutPose2019,MultiDeepLabCut2022}.
These works build automated tools that generally incorporate a pose detector~\cite{DeepLabCut2018,CNNAnimal2019,DeepLabCutPose2019} and sometimes also an object tracker~\cite{DeepPoseKit2019,MultiDeepLabCut2022}. Users first need to manually label a set of frames (normally several hundreds). Then the models are trained on this set and can be subsequently deployed on new instances from the same animal species. Such pipelines need to be carried out for each animal species of interest, and are not applicable to cases where multiple animal species are present. Other works train a single foundation model on datasets containing multiple animal species to achieve cross-species generalization~\cite{ViTPose2022,SuperAnimal2024}. They, however, manifest unsatisfying generalization performance.

We argue that instead of following this traditional paradigm and attempting to achieve generalized performance across animal species and scenarios, it is more desirable and practical to adopt a \textit{test time optimization} strategy. Specifically, we propose to train a model on each test instance. The model is trained on a sparse set of annotated frames from the input video, which can be obtained from manual annotation or a pre-trained pose detector~\cite{ViTPose2022}. The trained model is then applied to the rest of the frames. We build our model based on the crucial insight that we can share the knowledge of temporal tracking across all instances, while the unique knowledge we need in each example lies in appearance. Hence, we initialize our model from a state-of-the-art general-purpose point tracker~\cite{karaev2024cotracker3} and only fine-tune a lightweight \textit{appearance embedding} in it which encodes the appearance-specific information. With this strategy, our method achieves a 4-pixel accuracy of 81.6\% when supervised with only 6 annotated frames for a video of 60 frames, and the optimization converges in about 3mins\footnote{Correct when error is within 4 pixels. Averaged values over 15 videos.}. These values can be traded off depending on a preference for low annotation cost or high estimation quality.

Since our method does not rely on any particular assumption about animal species or morphology, it can be directly applied to any video of interest. We test our method on two datasets from different domains, namely DAVIS-Animals for quadruped animals and DeepFly3D~\cite{DeepFly3D2019} for tethered drosophilas. Our method produces high-quality per-frame annotations while achieving state-of-the-art performance on both datasets.

In summary, we propose a framework for dense animal pose annotation that follows a \textit{test time optimization} paradigm. We initialize the model with a general-purpose point tracker, and fine-tune a lightweight appearance embedding for each test instance. Our method can annotate cross-species animal videos at a high quality and can be scaled up at a reasonable cost. 
We evaluate our method on datasets covering different animal species and scenarios, and show that it produces state-of-the-art results. Check the project webpage for video results.


\begin{figure}
    \centering
    \includegraphics[width=\linewidth]{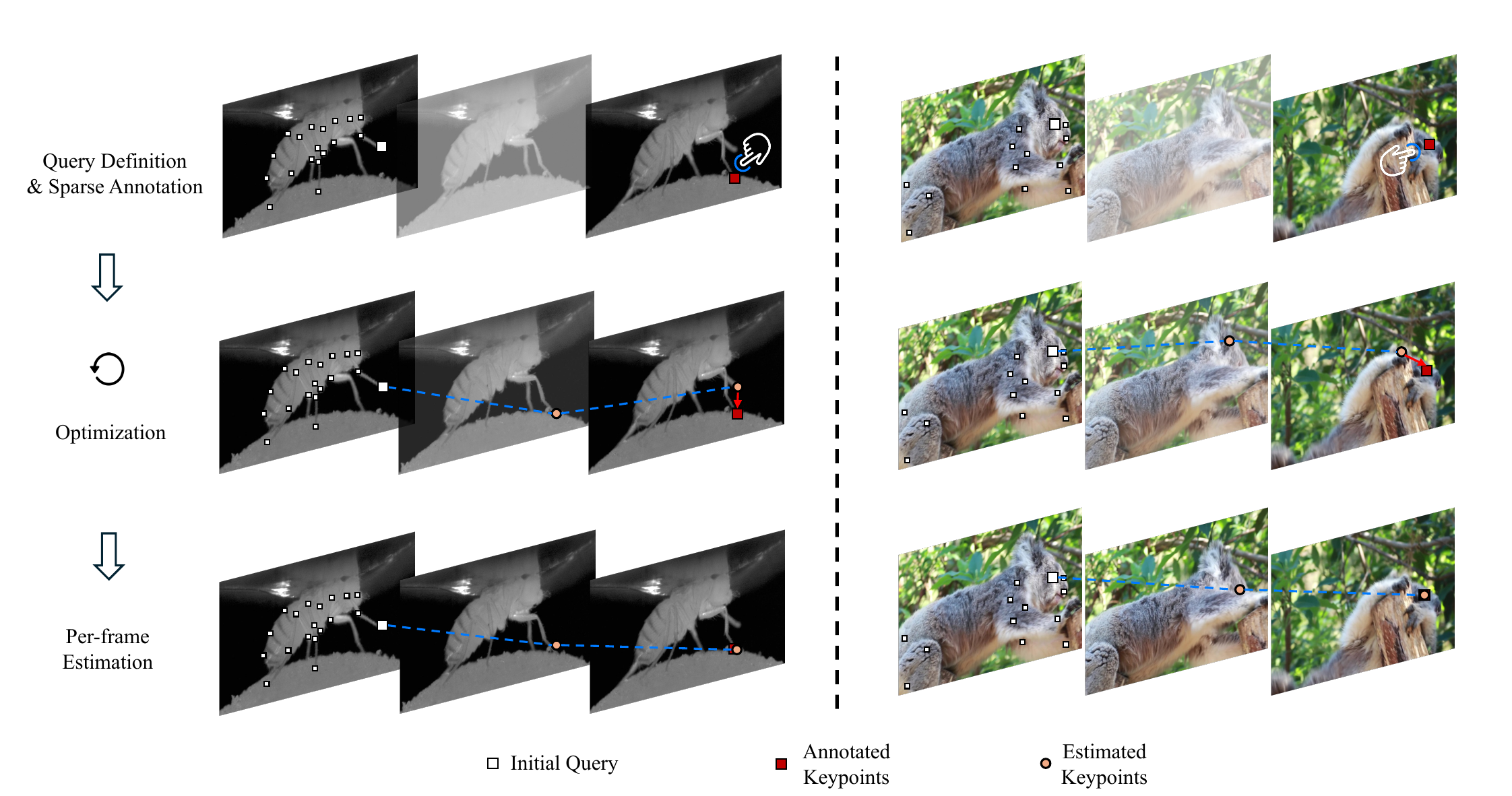}
    \caption{
    \textbf{System overview.} Our pipeline consists of three stages. First, users define query keypoints and provide sparse annotations. Our model is then optimized \wrt these annotations. Finally, the optimized model is applied to the remaining frames for dense pose labeling. We show two examples from the DeepFly3D (left) and DAVIS-Animals (right) datasets.
    }
    \vspace{-3mm}
    \label{fig:system}
\end{figure}
\section{Method}
\label{sec:method}

An overview of our pipeline is shown in~\cref{fig:system}. Our pipeline takes an input video sequence $\{\mathcal{I}_t\}_{t=1}^T$, query points $\{(\mathbf{x}^q_0, \mathbf{y}^q_0)\}$ on the first frame that define the keypoints to track, and a sparse set of annotated future keypoint positions $\{(\mathbf{x}^a_t, \mathbf{y}^a_t)\}_{t \in T_a}$, where $T_a$ represents the time steps at which the keypoints are annotated. 
The goal is to predict the keypoint positions in the remaining frames by leveraging general-purpose point trackers. We next provide essential background on a state-of-the-art point tracker CoTracker3~\cite{karaev2023cotracker}, upon which we build our pipeline. After that, we describe our test time optimization strategy in detail.

\subsection{Preliminaries: CoTracker3}
CoTracker3~\cite{karaev2024cotracker3} takes a video sequence $\{\mathcal{I}_t\}_{t=1}^T$ and query points $\{(\mathbf{x}^a_t, \mathbf{y}^a_t)\}$ as input to generate estimated point positions in each frame. 

\parahead{Feature encoding} 
CoTracker3 first computes dense feature maps with a convolutional neural network for each video frame, \ie $\Phi_t = \Phi(\mathcal{I}_t), t = 1, . . . , T$. The network downsamples the input video by a spatial factor $k=4$ and computes feature maps at $S=4$ different scales, \ie $\Phi_t^s \in \mathbb{R}^{d \times \frac{H}{k2^{s-1}} \times \frac{W}{k2^{s-1}}}, \quad s = 1, \dots, S.$ 

\parahead{Tracking features}
Points in CoTracker3 are described by extracting a square neighborhood with size $(2\Delta + 1)^2$ of features at different scales, \ie
\resizebox{\linewidth}{!}{
    $\phi_t^s = \left[ \Phi_t^s \left( \frac{\mathbf{x}_t}{ks} + \delta, \frac{\mathbf{y}_t}{ks} + \delta \right) : \delta \in \mathbb{Z}, \|\delta\|_{\infty} \leq \Delta \right], s = 1, \dots, S$
}. 
$\Phi_t^s(\mathbf{x}_t, \mathbf{y}_t)$ denotes binearly interpolating $\Phi_t^s$ around $(\mathbf{x}_t, \mathbf{y}_t)$, which can be either the query points or the current track estimates.

\parahead{Iterative updates}
Inference is carried out as an iterative update process.
The track estimates are first initialized from the positions of the query points. 
Each iteration begins by measuring the similarity between tracking features for current estimates $\phi_t^s$ and query points $\phi_{0}^s$, represented as correlation features $\text{Corr}_t = \text{MLP} \left( \langle \phi_{0}^s, \phi_t^s \rangle \right), s = 1, \dots, S$, where $\langle \phi_{0}^s, \phi_t^s \rangle = \text{stack}((\phi_0^s)^{\intercal} \phi_t^s)$. These correlation features along with other information are subsequently passed into a transformer to predict the position deltas. We refer to the original paper for all details.

\parahead{Discussions}
We find in our experiments that CoTracker3 does not perform reliably when applied out of the box to animal videos, possibly due to the scarcity of high-quality annotations across multiple animal species. It often produces stationary or highly jittery point estimates because it confuses the keypoints to track with points on background or other objects, which makes it unsuitable to be used in animal behavioral studies. In the following, we propose a test time optimization strategy that finetunes CoTracker3 on each test instance. This strategy greatly improves the pose detection performance at a manageable cost.

\begin{figure}
    \centering
    \includegraphics[width=\linewidth]{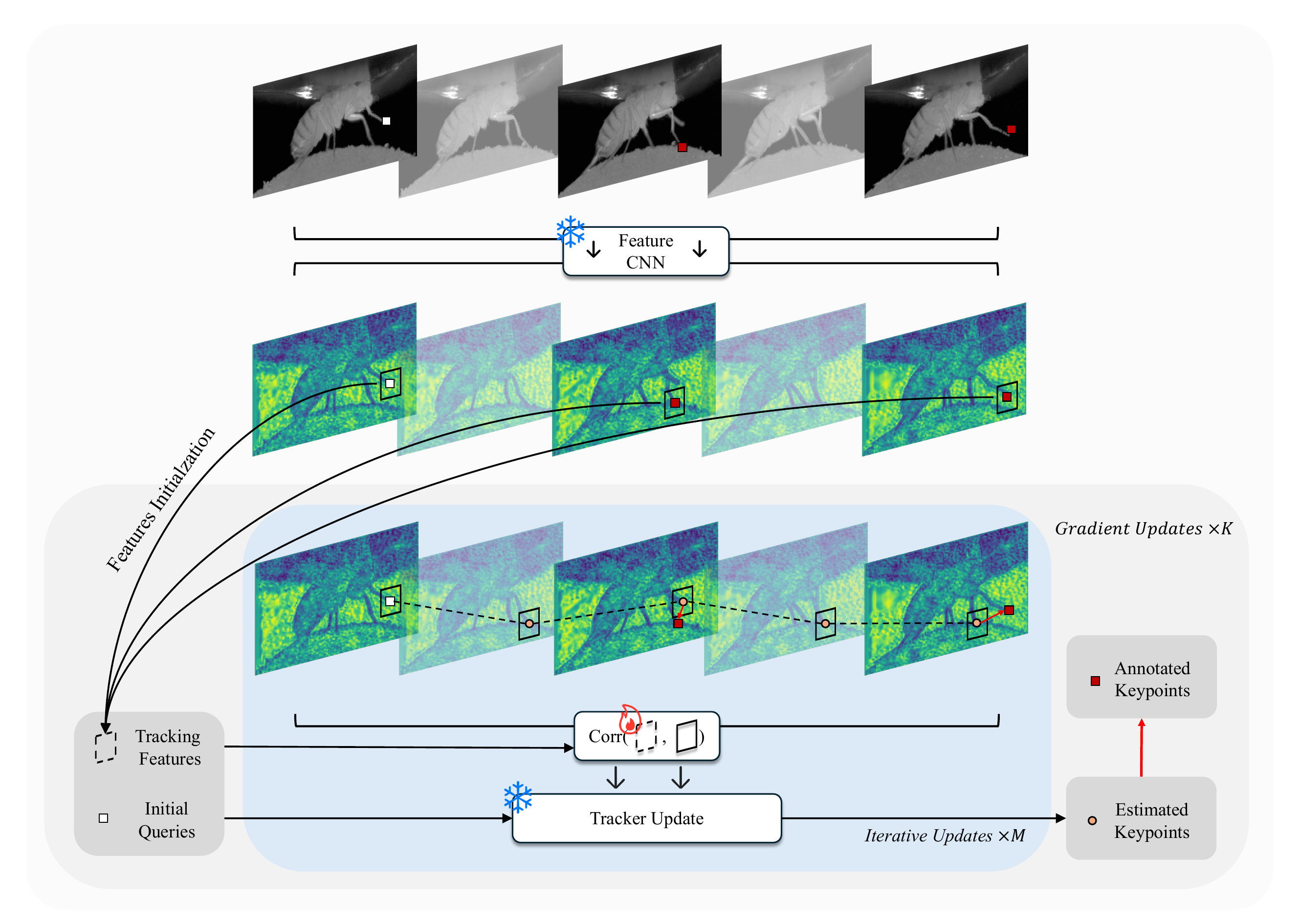}
    \caption{
    \textbf{Test time optimization.} 
    }
    \vspace{-3mm}
    \label{fig:method}
\end{figure}
\subsection{Test Time Optimization}
Our key idea is to optimize the tracking features for each query point $\hat{\phi}_0$ on a single video while keeping other components fixed. These optimized features essentially serve as an ``appearance embedding" that encodes video-specific appearance information, hence greatly helping improve tracking performance. We describe this process in detail below, which is also illustrated in~\cref{fig:method}.

\parahead{Feature initialization}
Instead of initializing the query features only from the square neighborhood of the first frame's feature map $\phi_0$, we choose to initialize with the square neighborhood of features from all the annotated frames $t \in T_a$:
\begin{equation}
    \hat{\phi}_0 := \frac{1}{|T_a|+1} \left(\phi_0^1 + \sum_{t \in T_a} \phi_t^1 \right)
\end{equation}

\parahead{Iterative updates}
The updated correlation features are calculated by
\begin{equation}
    \widehat{\text{Corr}}_t = \left( \text{MLP} \left( \langle \hat{\phi_{0}}, \phi_t^1 \rangle \right), \dots, \text{MLP} \left( \langle \hat{\phi_{0}}, \phi_t^S \rangle \right) \right)
\end{equation}
We then repeat the iterative update process as done in the original CoTracker3, using the updated correlation features.

\parahead{Loss function}
Our loss function consists of a primary tracking loss followed by a regularization term. For the tracking loss, we follow CoTracker3~\cite{karaev2024cotracker3} to supervise both the visible and occluded tracks using the Huber loss with a threshold of 6 and exponentially increasing weights:

\begin{equation}
   \mathcal{L}_{\text{track}} (\mathcal{P}, \mathcal{P}^{\star}) = \sum_{m=1}^M \gamma^{M-m} L_H(\mathcal{P}, \mathcal{P}^{\star}) 
\end{equation}
where $\mathcal{P}$ and $\mathcal{P^{\star}}$ denote the annotated and estimated keypoints, respectively. $M$ is the number of update iterations, $\gamma = 0.8$ is a discount factor, and $L_H$ is the Huber Loss. 

To ensure the tracking features remain close to the original sampled features, we apply an additional regularization term that penalizes significant deviations:

\begin{equation}
    \mathcal{L}_{\text{reg}} = \frac{1}{N} \sum_{i=1}^N\lVert \phi_0^{1i} - \hat{\phi}_0^{i} \rVert_1
\end{equation}
where $N$ is the number of keypoints. Our final loss is thus:

\begin{equation}
   \mathcal{L} = \mathcal{L}_{\text{track}} + \lambda \mathcal{L}_{\text{reg}} 
\end{equation}
We use $\lambda = 0.01$ in all experiments.

\parahead{Optimization details}
For each video, we optimize for 1,000 gradient update steps, starting with a learning rate of $1 \times 10^{-3}$ that linearly decays to $1 \times 10^{-5}$ using the Adam optimizer. For a 100-frame 480p video, the optimization converges in about 3 minutes on an NVIDIA RTX A6000.

\section{Experiments}

\begin{figure}
    \centering
    \includegraphics[width=\linewidth]{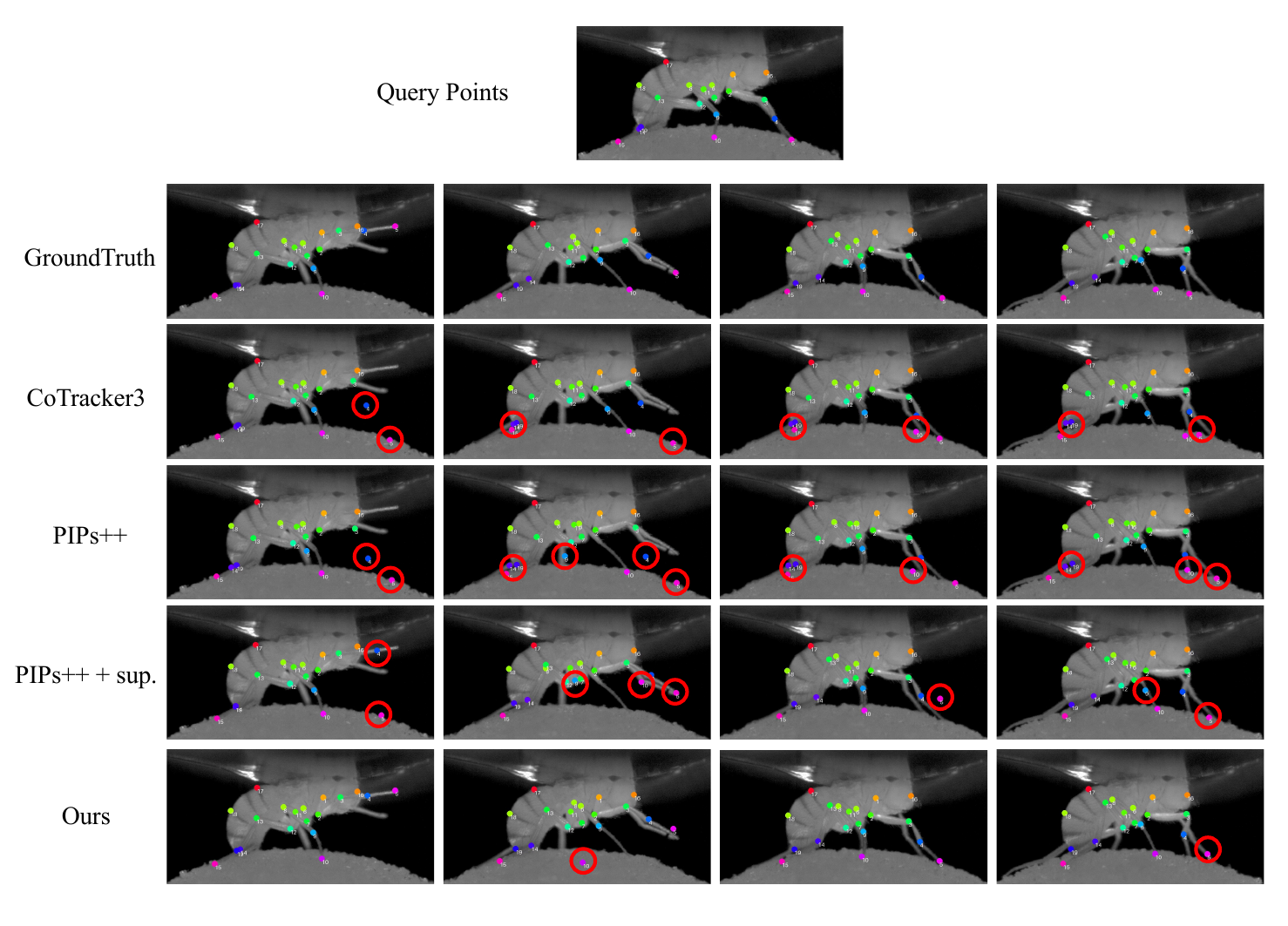}
    \caption{
    \textbf{Qualitative comparison on DeepFly3D.} Query points are shown in the frame on top, while estimated points in other frames. Keypoints with the same color / number denote correspondence. Red circles highlight estimation mistakes.
    }
    \label{fig:gallery-1}
\end{figure}
\begin{figure}
    \centering
     \includegraphics[width=\linewidth]{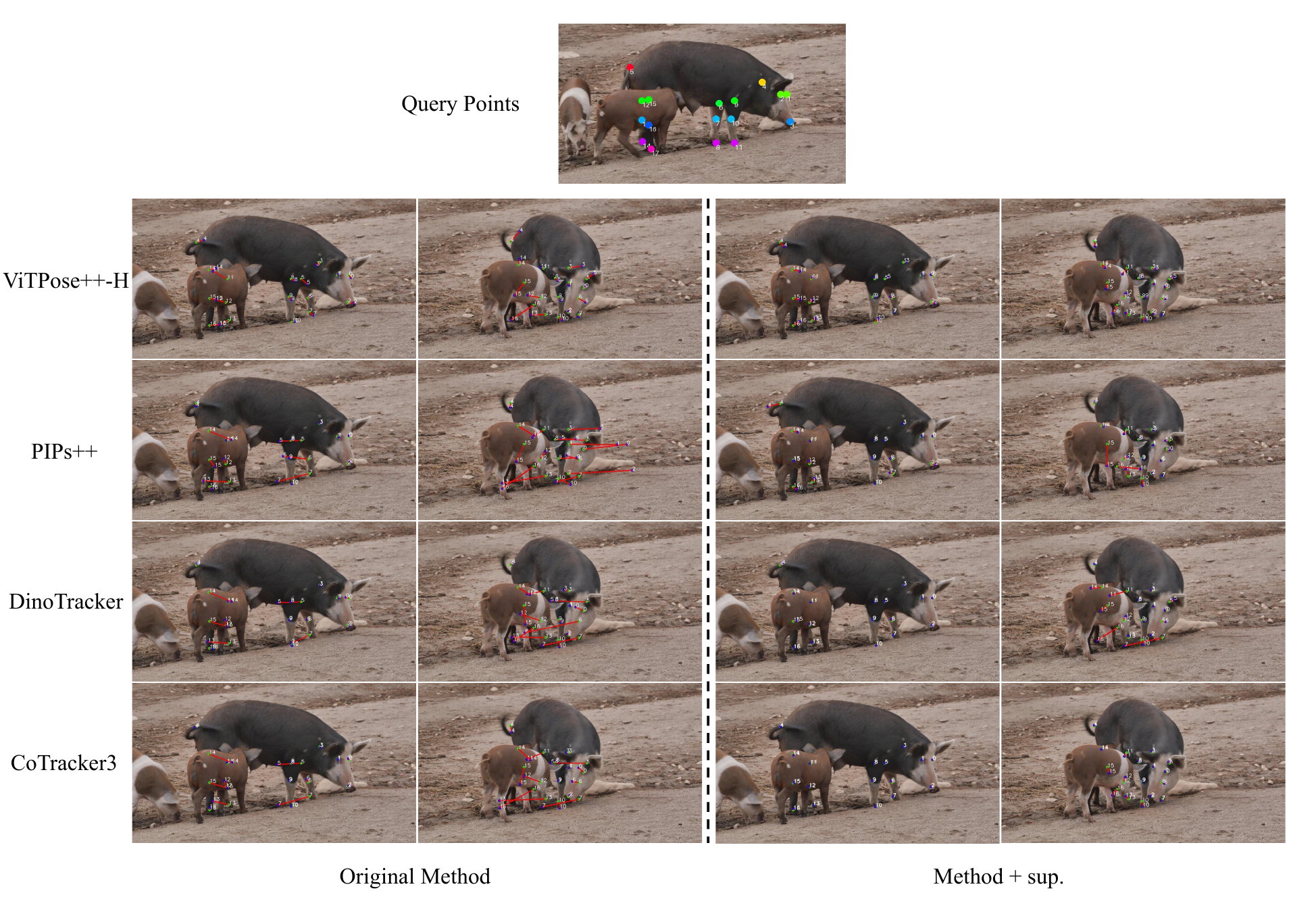}
    \caption{
    \textbf{Qualitative comparison on DAVIS-Animals.} Query points are color-coded in the frame on top. In other frames, estimated points are shown in blue, while ground-truth points are in green. Red lines indicate tracking errors relative to the corresponding ground truth positions. Note that CoTracker3 + sup. (bottom right) is our adapted version.
    }
    \label{fig:gallery-2}
\end{figure}

\parahead{Datasets} 
Animal videos with dense high-quality pose annotations are scarce. APT36k~\cite{APT36k2020} consists of 36k annotated frames from 2.4k videos of 30 animal species, but the keypoint annotations are very jittery. BADJA~\cite{BADJA2018} samples 7 animal videos from the DAVIS~\cite{DAVIS2017} dataset along with 2 other videos and provides high-quality dense annotations. However, the keypoint definition of BADJA differs from that of ViTPose~\cite{ViTPose2022}, which is an important baseline to compare. Hence, we sample 15 videos from DAVIS and manually annotate them, following the keypoint definition of ViTPose. We refer to this dataset as \textit{DAVIS-Animals}.

To demonstrate that our pipeline can be directly applied to videos of different animal species, we further evaluate our method on \textit{DeepFly3D}~\cite{DeepFly3D2019}, a dataset specifically for tethered drosophilas. We sample 15 videos from the dataset, each for 100 frames with 19 annotated keypoints.

\parahead{Metrics}
We follow SuperAnimal~\cite{SuperAnimal2024} to report 1) $\delta_{avg}$, which measures the average position accuracy of all points across 5 thresholds: 1, 2, 4, 8, and 16 pixels. Following~\cite{PointOdyssey2023}, images are resized to 256 × 256 pixels before calculation;
2) a jittering metric $J$, which is the average of the unsigned speed across all examples and keypoints. For a given keypoint $k$ and example $e$, $J_{k,e}$ is computed as:
\begin{equation}
    J_{k,e} = \frac{1}{N_{k,e}} \sum_{i=1}^{N_{k,e}} |v_{k,e,i}|
\end{equation}
where $N_{k,e}$ is the total number of speed measurements for keypoint $k$ in example $e$.
3) a masked jittering metric \( J_{\text{masked}} \), which enhances the jittering metric by focusing on correctly localized keypoints while penalizing incorrect ones. The metric is computed as:
\begin{equation}
J_{\text{masked},k,e} = \frac{1}{N_{k,e}} \sum_{i=1}^{N_{k,e}} v_{k,e,i} \cdot 
\begin{cases}
1, & d_{k,e,i} < 4 \\
10, & d_{k,e,i} \geq 4
\end{cases}
\end{equation}
where \( d_{k,e,i} \) is the distance to the ground truth. 

\parahead{Baselines}
We compare against state-of-the-art general-purpose point trackers (\textit{PIPS++}~\cite{PointOdyssey2023}, \textit{DINO-Tracker}~\cite{tumanyan2025dino}, and \textit{CoTracker3}~\cite{karaev2024cotracker3}), a cross-species pose estimator developed specifically for quadrupeds (\textit{SuperAnimal}~\cite{SuperAnimal2024}), as well as a general-purpose pose estimator applicable for both humans and animals(\textit{ViTPose}~\cite{ViTPose2022}). For a fair comparison, we also fine-tune them using the same strategy as our method (\textit{+sup.}). We next describe each in more detail.

\noindent\textit{PIPs++ (+sup.)} Similar to CoTracker3~\cite{karaev2024cotracker3}, \textit{PIPs++}~\cite{PointOdyssey2023} computes the cross correlation between the features at current timestep and that at previous timestep. For \textit{PIPS++ +sup.}, we instead compute the correlations between the current features and an optimizable features. 

\noindent\textit{DINO-Tracker (+sup.)} For each video, DINO-Tracker~\cite{tumanyan2025dino} fine-tunes a Delta-DINO encoder and a CNN-refiner through optical flows and DINO-feature correspondence. At inference time, they calculate the similarity between the refined DINO features sampled at the query point. Since this method is self-supervised, for a fair comparison, we add our tracking loss in addition to their original loss designs and finetune their original networks. 

\noindent\textit{ViTPose++-H (+sup.)} ViTPose++-H~\cite{ViTPose2022} is a pose estimation model trained on multiple humans and animals dataset. It is the SOTA model for animal pose estimation. We finetune this model on the annotated frames and test the performance of the fine-tuned model on the testing frames. 

\noindent\textit{SuperAnimal (+sup.)} We use the SuperAnimal-Quadruped model provided by Superanimal~\cite{SuperAnimal2024}, which is trained on multiple quadruped datasets. We enable their video-adaption option, which is another test time optimization technique they propose to improve estimation accuracy and mitigate jittering. They first obtain the pseudolabels every 10 frames by estimating the poses, and then use these estimated poses to finetune the estimator. We change their pseudolabels to the ground truth annotations and use these ground truth labels to finetune the estimator.
\begin{table}
\centering
\resizebox{\linewidth}{!}{ 
\begin{tabular}{lccccccc}
\toprule
\multirow{2}{*}{\textbf{Method}} & \multicolumn{3}{c}{\textbf{DeepFly3D}} & \multicolumn{3}{c}{\textbf{DAVIS-Animals}} \\
\cmidrule(lr){2-4}
\cmidrule(lr){5-7}
& $\delta_{avg}\uparrow$ & $J\downarrow$ & $J_{masked}\downarrow$ & $\delta_{avg}\uparrow$ & $J\downarrow$ & $J_{masked}\downarrow$\\
\midrule
SuperAnimal~\cite{SuperAnimal2024} & --- & --- & --- & 41.07  & 15.19 & 10.90 \\
ViTPose++-H~\cite{ViTPose2022} & --- & --- & --- & 52.56 & 9.52 & 9.22 \\
PIPs++~\cite{PointOdyssey2023}& 45.91 & \textbf{0.55} & 5.32 & 46.19 & \textbf{4.41} & 7.67 \\
DINO-Tracker~\cite{tumanyan2025dino} & 48.51 & 1.95 & 5.40 & 49.09 & 8.41 & 8.61 \\
CoTracker3~\cite{karaev2024cotracker3} & 50.26 & 0.60 & 4.92 & 49.59 & 6.26 & 8.03 \\
\midrule
SuperAnimal +sup. & --- & --- & --- & 57.54  & 10.36 & 9.19 \\
ViTPose++-H +sup. & --- & --- & --- & 62.17 & 8.21 & 7.94 \\
PIPs++ +sup. & 58.89 & 2.14 & 3.72 & 51.17 & 8.20 & 8.13 \\
DINO-Tracker+sup. & 67.29 & 2.91 & 3.39 & 63.33 & 9.11 & 8.67 \\
Ours & \textbf{70.80} & \textbf{1.46} & \textbf{2.54} & \textbf{67.53} & \textbf{7.04} & \textbf{7.15} \\
\bottomrule
\end{tabular}%
}
\vspace{-1mm}
\captionof{table}{ 
Quantitative comparison with baselines.
}
\vspace{-3mm}
\label{tab:comparison}
\end{table}
\subsection{Quantitative Comparison}

For this experiment, we train our method and baselines with additional supervision on frames 0, 10, ..., and test on frames 5, 15, .... The results are provided in~\cref{tab:comparison}. Our method achieves the highest $\delta_{avg}$ score on both datasets, surpassing baselines that are applied out of the box or fine-tuned with test time optimization (\textit{+sup.}). For the jittering metric, our method performs the best compared to baselines fine-tuned with test time optimization, but is worse than PIPs++ and CoTracker3. This is because PIPs++ and CoTracker3 produce estimates that confuse with the background more often, which tend to move less and result in a deceptively low jittering score (\eg keypoints 4 and 5 in~\cref{fig:gallery-1}). To better evaluate the tracking quality, we use the masked jittering metric $J_{masked}$, which focuses on correctly localized keypoints while penalizing incorrect ones. As shown in ~\cref{tab:comparison}, our method achieves the lowest $J_{masked}$ score across all methods.

\subsection{Qualitative Comparison}
We compare qualitatively with selected baselines in~\cref{fig:gallery-1,fig:gallery-2}. In~\cref{fig:gallery-1}, both CoTracker3~\cite{karaev2024cotracker3} and PIPs++~\cite{PointOdyssey2023} have obvious errors on keypoints on legs. For PIPs++, tracking is noticeably more accurate after test time optimization but errors still occur frequently. Comparatively, our method is able to track the keypoints most reliably and produce the least errors.
In~\cref{fig:gallery-2} we observe similar trends. For all methods, fine-tuning with test time optimization noticeably reduces tracking errors. Among all methods, our method is the most accurate, also producing the least jitter.
\section{Conclusion}
We propose an animal pose labeling pipeline that features high-quality pose labeling while at a reasonable cost for manual annotation. Our key idea is to adopt a test time optimization strategy to fine-tune a pretrained model with sparse annotations on the example we wish to annotate. By leveraging a general-purpose point tracker, we inherit valuable knowledge of temporal tracking and only fine-tune a lightweight appearance embedding that encodes example-specific appearance information. Our method achieves state-of-the-art pose estimation performance across different animal species, offering a valuable tool for accurate animal behavior quantification.

{
    \small
    \bibliographystyle{ieeenat_fullname}
    \bibliography{main}

\begin{thebibliography}{18}
\providecommand{\natexlab}[1]{#1}
\providecommand{\url}[1]{\texttt{#1}}
\expandafter\ifx\csname urlstyle\endcsname\relax
  \providecommand{\doi}[1]{doi: #1}\else
  \providecommand{\doi}{doi: \begingroup \urlstyle{rm}\Url}\fi

\bibitem[Biggs et~al.(2018)Biggs, Roddick, Fitzgibbon, and Cipolla]{BADJA2018}
Benjamin Biggs, Thomas Roddick, Andrew Fitzgibbon, and Roberto Cipolla.
\newblock {C}reatures great and {SMAL}: {R}ecovering the shape and motion of animals from video.
\newblock In \emph{ACCV}, 2018.

\bibitem[{Cao} et~al.(2019){Cao}, {Hidalgo Martinez}, {Simon}, {Wei}, and {Sheikh}]{OpenPose2019}
Z. {Cao}, G. {Hidalgo Martinez}, T. {Simon}, S. {Wei}, and Y.~A. {Sheikh}.
\newblock Openpose: Realtime multi-person 2d pose estimation using part affinity fields.
\newblock \emph{IEEE Transactions on Pattern Analysis and Machine Intelligence}, 2019.

\bibitem[Doersch et~al.(2022)Doersch, Gupta, Markeeva, Recasens, Smaira, Aytar, Carreira, Zisserman, and Yang]{doersch2022tap}
Carl Doersch, Ankush Gupta, Larisa Markeeva, Adria Recasens, Lucas Smaira, Yusuf Aytar, Joao Carreira, Andrew Zisserman, and Yi Yang.
\newblock Tap-vid: A benchmark for tracking any point in a video.
\newblock \emph{Advances in Neural Information Processing Systems}, 35:\penalty0 13610--13626, 2022.

\bibitem[Goel et~al.(2023)Goel, Pavlakos, Rajasegaran, Kanazawa, and Malik]{HMR22023}
Shubham Goel, Georgios Pavlakos, Jathushan Rajasegaran, Angjoo Kanazawa, and Jitendra Malik.
\newblock Humans in 4d: Reconstructing and tracking humans with transformers.
\newblock In \emph{Proceedings of the IEEE/CVF International Conference on Computer Vision}, pages 14783--14794, 2023.

\bibitem[Graving et~al.(2019)Graving, Chae, Naik, Li, Koger, Costelloe, and Couzin]{DeepPoseKit2019}
Jacob~M Graving, Daniel Chae, Hemal Naik, Liang Li, Benjamin Koger, Blair~R Costelloe, and Iain~D Couzin.
\newblock Deepposekit, a software toolkit for fast and robust animal pose estimation using deep learning.
\newblock \emph{Elife}, 8:\penalty0 e47994, 2019.

\bibitem[G{\"u}nel et~al.(2019)G{\"u}nel, Rhodin, Morales, Campagnolo, Ramdya, and Fua]{DeepFly3D2019}
Semih G{\"u}nel, Helge Rhodin, Daniel Morales, Jo{\~a}o Campagnolo, Pavan Ramdya, and Pascal Fua.
\newblock Deepfly3d, a deep learning-based approach for 3d limb and appendage tracking in tethered, adult drosophila.
\newblock \emph{Elife}, 8:\penalty0 e48571, 2019.

\bibitem[Karaev et~al.(2023)Karaev, Rocco, Graham, Neverova, Vedaldi, and Rupprecht]{karaev2023cotracker}
Nikita Karaev, Ignacio Rocco, Benjamin Graham, Natalia Neverova, Andrea Vedaldi, and Christian Rupprecht.
\newblock Cotracker: It is better to track together.
\newblock \emph{arXiv preprint arXiv:2307.07635}, 2023.

\bibitem[Karaev et~al.(2024)Karaev, Makarov, Wang, Neverova, Vedaldi, and Rupprecht]{karaev2024cotracker3}
Nikita Karaev, Iurii Makarov, Jianyuan Wang, Natalia Neverova, Andrea Vedaldi, and Christian Rupprecht.
\newblock Cotracker3: Simpler and better point tracking by pseudo-labelling real videos.
\newblock \emph{arXiv preprint arXiv:2410.11831}, 2024.

\bibitem[Lauer et~al.(2022)Lauer, Zhou, Ye, Menegas, Schneider, Nath, Rahman, Di~Santo, Soberanes, Feng, et~al.]{MultiDeepLabCut2022}
Jessy Lauer, Mu Zhou, Shaokai Ye, William Menegas, Steffen Schneider, Tanmay Nath, Mohammed~Mostafizur Rahman, Valentina Di~Santo, Daniel Soberanes, Guoping Feng, et~al.
\newblock Multi-animal pose estimation, identification and tracking with deeplabcut.
\newblock \emph{Nature Methods}, 19\penalty0 (4):\penalty0 496--504, 2022.

\bibitem[Mathis et~al.(2018)Mathis, Mamidanna, Cury, Abe, Murthy, Mathis, and Bethge]{DeepLabCut2018}
Alexander Mathis, Pranav Mamidanna, Kevin~M Cury, Taiga Abe, Venkatesh~N Murthy, Mackenzie~Weygandt Mathis, and Matthias Bethge.
\newblock Deeplabcut: markerless pose estimation of user-defined body parts with deep learning.
\newblock \emph{Nature neuroscience}, 21\penalty0 (9):\penalty0 1281--1289, 2018.

\bibitem[Nath et~al.(2019)Nath, Mathis, Chen, Patel, Bethge, and Mathis]{DeepLabCutPose2019}
Tanmay Nath, Alexander Mathis, An~Chi Chen, Amir Patel, Matthias Bethge, and Mackenzie~Weygandt Mathis.
\newblock Using deeplabcut for 3d markerless pose estimation across species and behaviors.
\newblock \emph{Nature protocols}, 14\penalty0 (7):\penalty0 2152--2176, 2019.

\bibitem[Pereira et~al.(2019)Pereira, Aldarondo, Willmore, Kislin, Wang, Murthy, and Shaevitz]{CNNAnimal2019}
Talmo~D Pereira, Diego~E Aldarondo, Lindsay Willmore, Mikhail Kislin, Samuel S-H Wang, Mala Murthy, and Joshua~W Shaevitz.
\newblock Fast animal pose estimation using deep neural networks.
\newblock \emph{Nature methods}, 16\penalty0 (1):\penalty0 117--125, 2019.

\bibitem[Pont-Tuset et~al.(2017)Pont-Tuset, Perazzi, Caelles, Arbel{\'a}ez, Sorkine-Hornung, and Van~Gool]{DAVIS2017}
Jordi Pont-Tuset, Federico Perazzi, Sergi Caelles, Pablo Arbel{\'a}ez, Alex Sorkine-Hornung, and Luc Van~Gool.
\newblock The 2017 davis challenge on video object segmentation.
\newblock \emph{arXiv preprint arXiv:1704.00675}, 2017.

\bibitem[Tumanyan et~al.(2025)Tumanyan, Singer, Bagon, and Dekel]{tumanyan2025dino}
Narek Tumanyan, Assaf Singer, Shai Bagon, and Tali Dekel.
\newblock Dino-tracker: Taming dino for self-supervised point tracking in a single video.
\newblock In \emph{European Conference on Computer Vision}, pages 367--385. Springer, 2025.

\bibitem[Xu et~al.(2022)Xu, Zhang, Zhang, and Tao]{ViTPose2022}
Yufei Xu, Jing Zhang, Qiming Zhang, and Dacheng Tao.
\newblock Vitpose: Simple vision transformer baselines for human pose estimation.
\newblock \emph{Advances in Neural Information Processing Systems}, 35:\penalty0 38571--38584, 2022.

\bibitem[Yang et~al.(2022)Yang, Yang, Xu, Zhang, Lan, and Tao]{APT36k2020}
Yuxiang Yang, Junjie Yang, Yufei Xu, Jing Zhang, Long Lan, and Dacheng Tao.
\newblock Apt-36k: A large-scale benchmark for animal pose estimation and tracking.
\newblock \emph{Advances in Neural Information Processing Systems}, 35:\penalty0 17301--17313, 2022.

\bibitem[Ye et~al.(2024)Ye, Filippova, Lauer, Schneider, Vidal, Qiu, Mathis, and Mathis]{SuperAnimal2024}
Shaokai Ye, Anastasiia Filippova, Jessy Lauer, Steffen Schneider, Maxime Vidal, Tian Qiu, Alexander Mathis, and Mackenzie~Weygandt Mathis.
\newblock Superanimal pretrained pose estimation models for behavioral analysis.
\newblock \emph{Nature Communications}, 15\penalty0 (1):\penalty0 5165, 2024.

\bibitem[Zheng et~al.(2023)Zheng, Harley, Shen, Wetzstein, and Guibas]{PointOdyssey2023}
Yang Zheng, Adam~W Harley, Bokui Shen, Gordon Wetzstein, and Leonidas~J Guibas.
\newblock Pointodyssey: A large-scale synthetic dataset for long-term point tracking.
\newblock In \emph{Proceedings of the IEEE/CVF International Conference on Computer Vision}, pages 19855--19865, 2023.

\end{thebibliography}
}

\appendix
\clearpage
\setcounter{page}{1}
\maketitlesupplementary

\section{Ablation Study}
\parahead{Fine-tuned components}
We experiment with fine-tuning different components of CoTracker3. In addition to fine-tuning the query point embedding (\textit{qp\_emb}), we also try fine-tuning the feature extractor CNN (\textit{FNet}) as well as fine-tuning all components (\textit{All}), which include qp\_emb, FNet, and the final transformer. We perform the ablation study on DAVIS-Animals and report the results in~\cref{tab:ablation}. Finetuning just the query point embedding proves to be the most effective, backed up by its highest $\delta_{avg}$, lowest $J$ (apart from original CoTracker3, for reasons explained above), and lowest $J_{masked}$ scores. It also converges in a much shorter time.

\begin{figure}
    \centering
    \includegraphics[width=\linewidth]{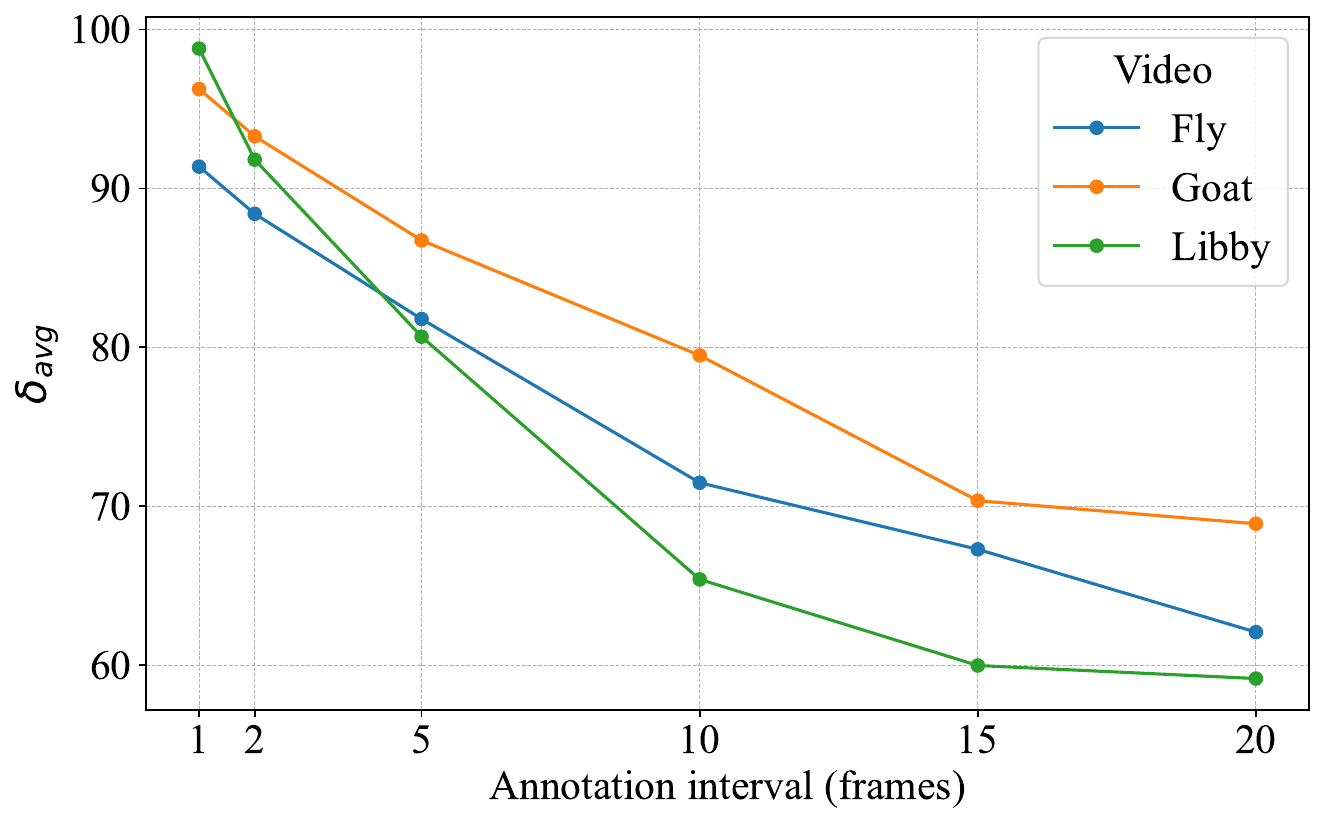}
    \vspace{-5mm}
    \caption{
        Effect of different numbers of annotated frames.
    }
    \vspace{-3mm}
    \label{fig:frames}
\end{figure}

\parahead{Number of annotated frames}
We study the effect of optimizing on different numbers of annotated frames. We select videos—Fly (100 frames), Goat (90 frames), and Libby (49 frames)—and provide annotations at different frame intervals (\eg, an interval of 1 means annotating every frame). The annotation intervals include 1/2/5/10/15/20 frames, resulting in 100/50/20/10/6/5 annotated frames for Fly, 90/45/18/9/6/4 for Goat, and 49/24/9/4/3/2 for Libby. We plot the relationship between the annotation interval and the $\delta_{avg}$ in~\cref{fig:frames}. As expected, we observe an increase in performance as more annotations are provided. 

\begin{figure}
    \centering
    \includegraphics[width=\linewidth]{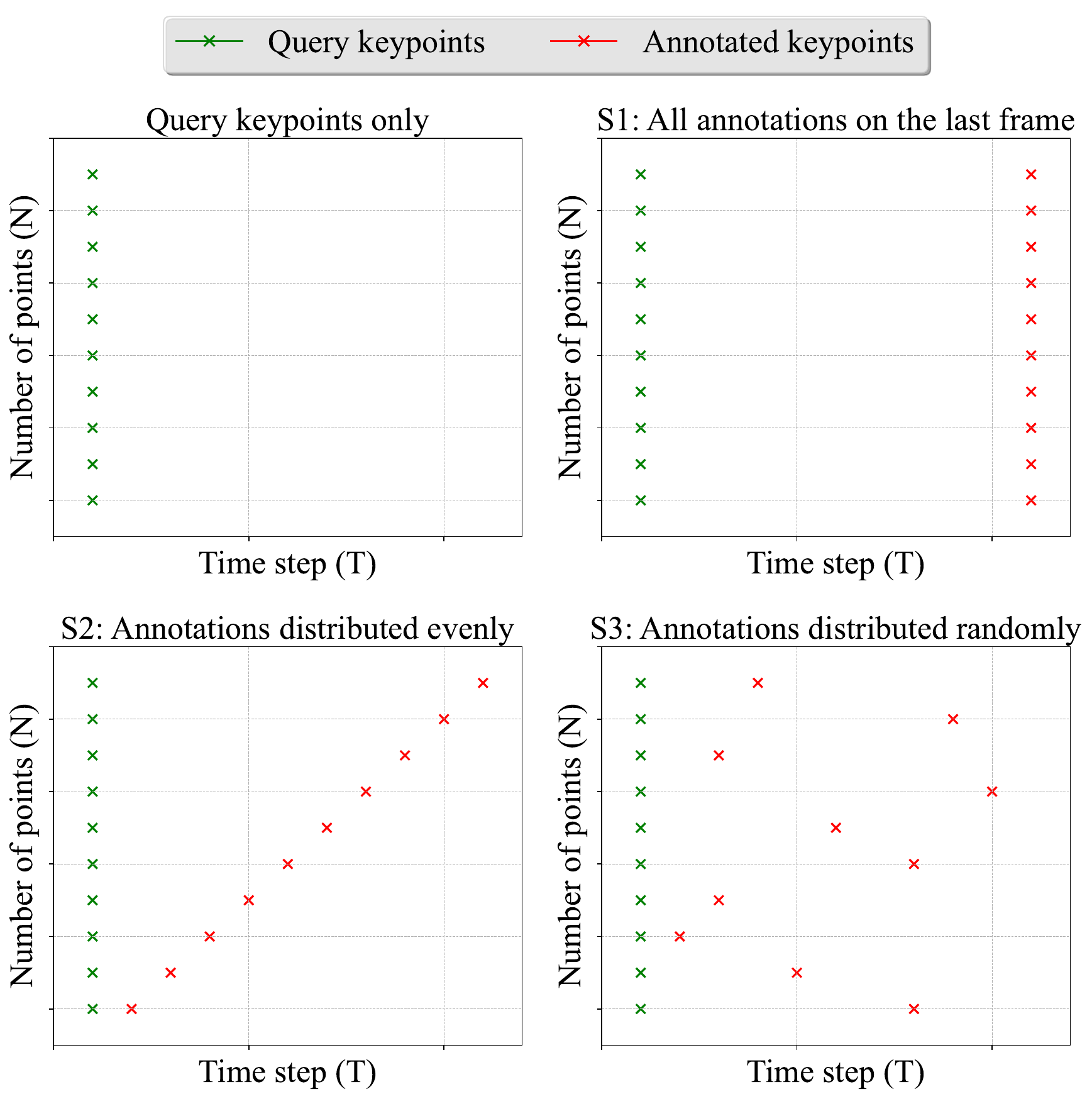}
    \caption{
        Illustration of different keypoint annotation strategies.
    }
    \label{fig:strategy}
\end{figure}

\parahead{Different strategies of annotating keypoints}
It seems most natural to annotate all keypoints in each annotated frame, which is also what we do throughout our experiments. However, we are interested in alternative annotation strategies. 
We investigate this on the DeepFly3D dataset. Each video has 100 frames and there are 19 keypoints of interest. Concretely, we experiment with three annotation strategies:
    \textbf{S1:} Annotate all keypoints in the last frame;
    \textbf{S2:} Annotate keypoints evenly distributed across frames 1-100;
    \textbf{S3:} Annotate keypoints randomly distributed across frames 1-100.
These strategies are also illustrated in~\cref{fig:strategy}. We report the results with them in~\cref{tab:strategy}. Among the three strategies, S2 performs the best. We hypothesize that diverse temporal cues improve model generalization.

\begin{table}
\centering
\resizebox{\linewidth}{!}{ 
\begin{tabular}{lccccc}
\toprule
\textbf{Method} & $\delta_{avg}\uparrow$ & $J\downarrow$ & $J_{masked}\downarrow$ & Time(min)$\downarrow$ \\
\midrule
CoTracker3~\cite{karaev2024cotracker3} & 49.59 & \textbf{6.26} & 8.03 & - \\+sup.(FNet) & 54.08 & 7.40 & 7.54 & 9.6 \\ 
+sup.(All) & 66.26 & 7.32 & 7.34 & 10.1 \\
+sup.(ours/qp\_emb) & \textbf{67.53} & 7.04 & \textbf{7.15} & \textbf{3.1} \\

\bottomrule
\end{tabular}%
}
\caption{
Ablation study on network components to fine-tune.
}
\vspace{-3mm}
\label{tab:ablation}
\end{table}
\begin{table}
\centering
\begin{tabular}{cc}
\toprule
\textbf{Annotation strategy} & $\delta_{avg}\uparrow$ \\
\midrule
S1 & 44.86 \\
S2 & \textbf{54.39} \\ 
S3 & 44.61 \\
\bottomrule
\end{tabular}%
\caption{
    Ablation study on keypoint annotation strategies.
}
\vspace{-3mm}
\label{tab:strategy}
\end{table}

\parahead{Pseudo-labels vs. manual annotations}
Our method performs best when it is supervised with manually annotated keypoints. However, this does incur additional annotation overhead which may not be desirable in certain applications (\eg those that prefer annotation efficiency over accuracy). We hence study the effect of supervising our model with pseudo-labels. For this experiment, we take the DAVIS-Animals dataset and use the ViTPose++-H~\cite{ViTPose2022} model to generate keypoint estimations on the same set of frames that are manually annotated, and train a model supervised with each annotation set. The results are summarized in~\cref{tab:annot}. As expected, the model supervised with manual annotations significantly outperforms its counterpart supervised with pseudo-labels. This offers a trade-off between annotation quality and efficiency, depending on the preference of the down-stream application.

\section{Videos with Long Duration and Fast Motion} 
While more accurate annotations generally improve performance, full video annotation is not strictly required. Our results show that even partial annotations yield better performance than using tracking models alone (\eg CoTracker3). To validate this, we sampled sequences from the DeepFly3D dataset at lengths of 100, 200, and 600 frames, annotating every 10th frame. Our method consistently outperforms Cotracker3, with full-sequence annotation providing better results than annotating only the first 100 frames.

\begin{figure}
    \centering
    \includegraphics[width=\linewidth]{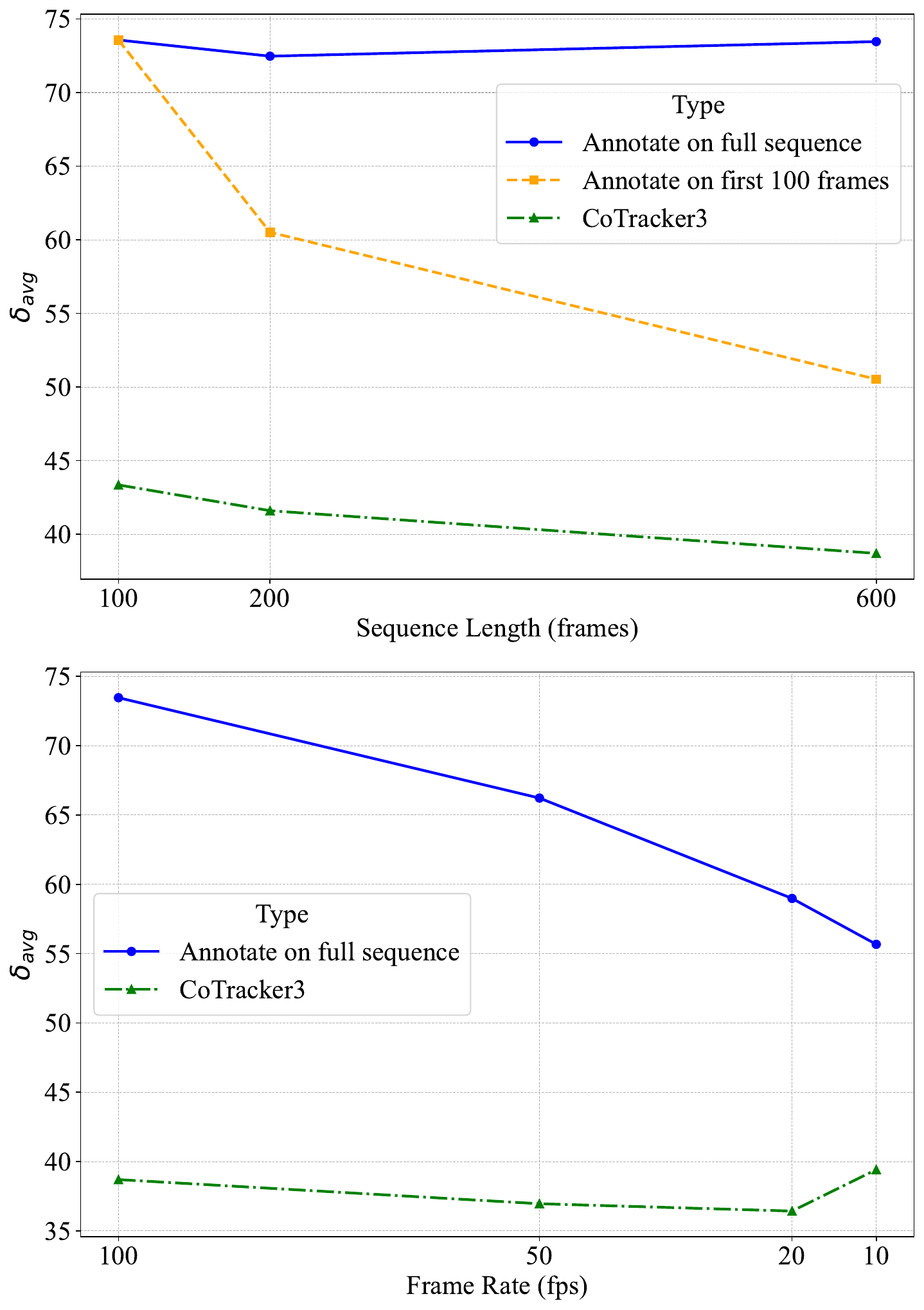}
    \caption{
        Long duration and fast movement experiments.
    }
    \vspace{-3mm}
    \label{fig:line_lf}

\end{figure}

 For fast movement, we used fly sequences (originally 100fps, sampled to 600 frames each) and downsampled to simulate effective frame rates of 50, 20, and 10fps, making the motion appear faster.  As shown in Figure \ref{fig:line_lf}, our method maintains over 55\% accuracy even at 10x speed, showing its robustness in challenging scenarios.
\section{Post-Processing with Kalman Filter}

One area where our method clearly outperforms baselines is that our estimation results have noticeably less jittering, as can be seen in the supplementary video. 
We hence perform an additional experiment to investigate whether applying the Kalman filter as post-processing directly to the estimation results can effectively reduce such artifacts. This is also used in SuperAnimal~\cite{SuperAnimal2024}.

In~\cref{tab:comparison_filter}, we report the results comparing original methods with adding additional Kalman filtering. We set the number of iterations to 10 for the filter. 
We find that using filtering as an additional post-processing step improves all methods on both the accuracy and jittering metrics, and our method outperforms all baselines when additional filtering is applied as well.
\begin{table}
\centering
\resizebox{\linewidth}{!}{ 
\begin{tabular}{lcccc}
\toprule
\textbf{Method} & $\delta_{avg}\uparrow$ & $J\downarrow$ & $J_{masked}\downarrow$ \\
\midrule
CoTracker3~\cite{karaev2024cotracker3} & 49.59 & \textbf{6.26} & 8.03\\
+sup.(w/ pseudo-labels) & 51.17 & 8.08 & 8.74\\ 
+sup.(w/ manual annotations) & \textbf{67.53} & 7.04 & \textbf{7.15}\\

\bottomrule
\end{tabular}%
}
\captionof{table}{
    Ablation study on using pseudo-ground-truth.
}
\vspace{-3mm}
\label{tab:annot}
\end{table}

\begin{table}
\centering
\begin{tabular}{lccccc}
\toprule
\multirow{2}{*}{\textbf{Method}} & \multicolumn{2}{c}{\textbf{DAVIS-Animals}} \\
\cmidrule(lr){2-3}
& $\delta_{avg}\uparrow$ & $J\downarrow$\\
\midrule
SuperAnimal +sup. & 57.54  & 10.36 \\
SuperAnimal +sup. +Kalman Filter & 59.07 & 7.57 \\
ViTPose++-H +sup. & 62.17 & 8.21 \\
ViTPose++-H +sup. +Kalman Filter & 62.40 & 7.76 \\
PIPs++ +sup. & 51.17 & 8.20 \\
PIPs++ +sup. +Kalman Filter & 51.52 & 7.97 \\
DINO-Tracker+sup. & 63.33 & 9.11\\
DINO-Tracker+sup. +Kalman Filter & 61.91 & 7.64\\
Ours & 67.53 & 7.04 \\
Ours +Kalman Filter & \textbf{67.82}  & \textbf{6.97} \\
\bottomrule
\end{tabular}%
\captionof{table}{ 
Comparison when using additional Kalman filtering.
}
\vspace{-5mm}
\label{tab:comparison_filter}
\end{table}

\begin{figure*}[ht]
    \centering
     \includegraphics[width=\linewidth]{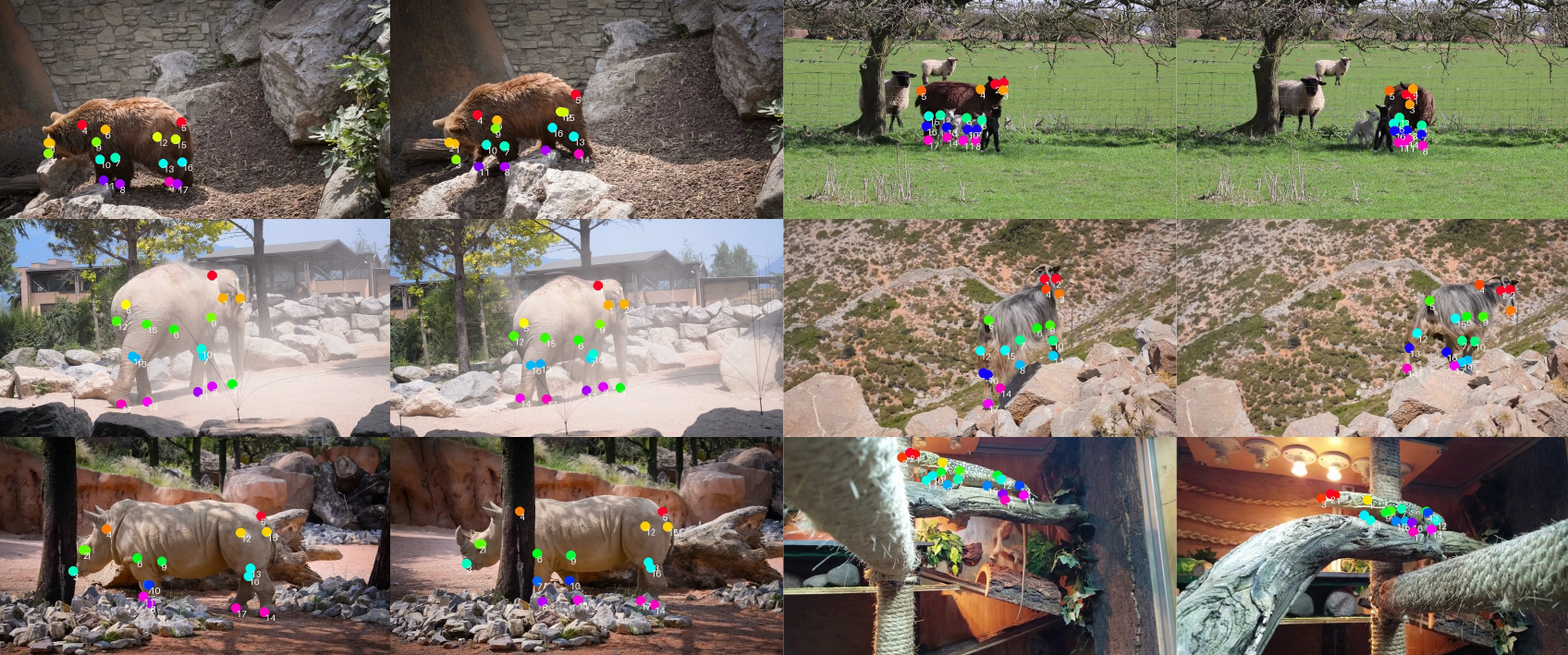}
    \caption{
    \textbf{More results on DAVIS-Animals.} We show 6 examples, each with two frames overlaid with estimated keypoints.
    }
    \label{fig:gallery-sup}
\end{figure*}
\begin{figure*}
    \centering
     \includegraphics[width=\linewidth]{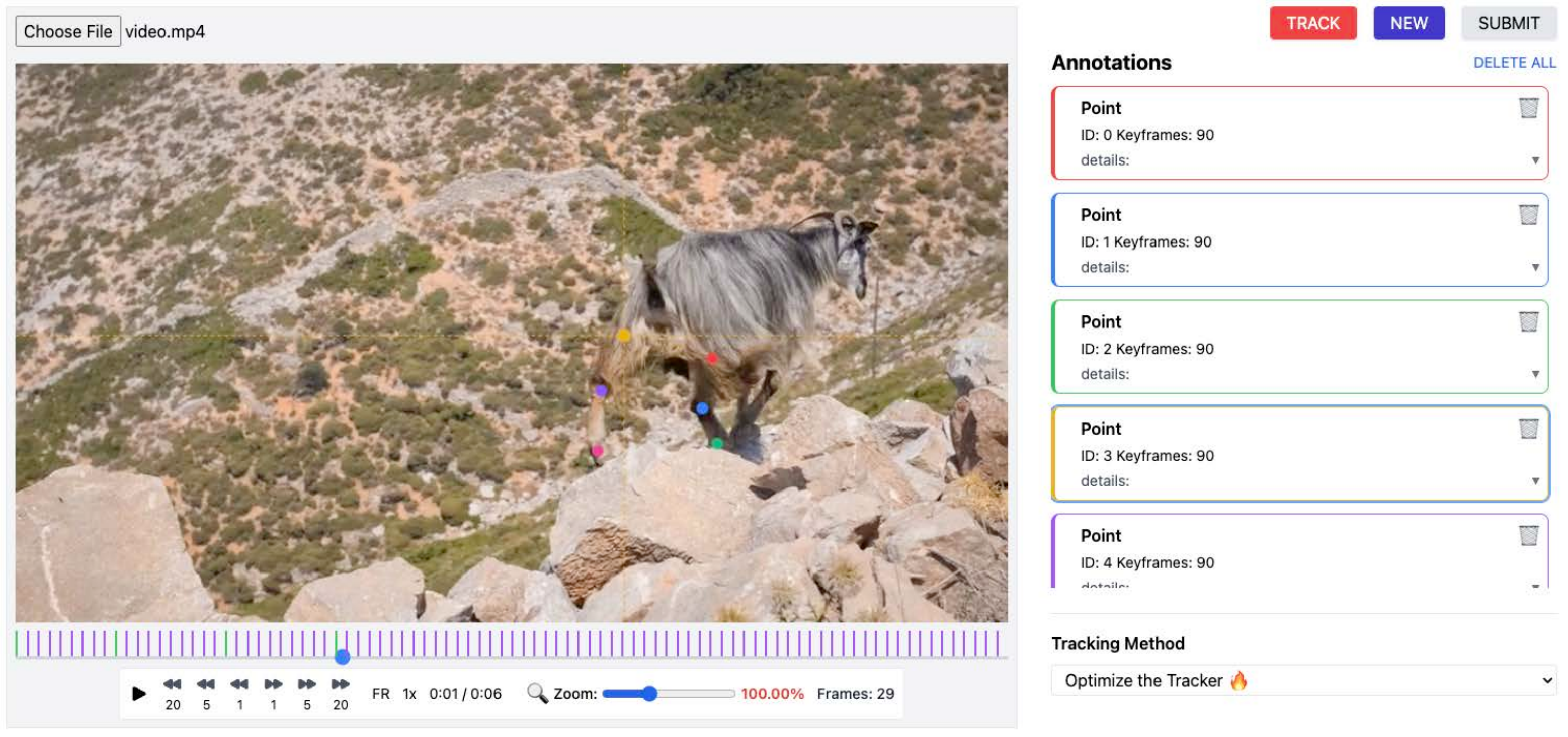}
    \caption{
    \textbf{Our Labeling Interface.} We adopt the interface design from~\cite{doersch2022tap}. Users can add manual annotations by simply clicking on the canvas, and leverage a tracker to assist with the annotation process.
    }
    \vspace{-3mm}
    \label{fig:gui}
\end{figure*}

\section{Labeling Interface}

We provide a labeling interface designed for efficient and user-friendly annotation (see Figure~\ref{fig:gui} and the supplementary video). The interface follows the design of Doersch et al.~\cite{doersch2022tap}, with interactive elements for both manual and assisted labeling.

To start, users upload the video they wish to annotate. They can then add keypoints by simply clicking on the canvas. Once the first frame is annotated, users can click the \texttt{TRACK} button to run the default CoTracker3~\cite{karaev2024cotracker3} tracker, which generates coarse label predictions across the video.

If any keypoints are misaligned or off-target, users can manually correct them by dragging the points to the correct locations. To refine the trajectories further, users can switch the tracking method to our optimization-based version using the dropdown menu, and then click the \texttt{TRACK} button again. This launches our optimization process, which improves temporal consistency and overall accuracy.

The video can then be replayed within the interface for users to inspect the results. If the annotations are satisfactory, the \texttt{SUBMIT} button allows users to finalize and download their labels. Additional controls such as frame-by-frame navigation, zoom, and playback speed adjustment are also provided to make the annotation experience smoother.

\section{Additional results}
We show additional results in ~\cref{fig:gallery-sup}.

\end{document}